\documentclass[letterpaper]{article} 
\usepackage{aaai25}  
\usepackage{times}  
\usepackage{helvet}  
\usepackage{courier}  
\usepackage[hyphens]{url}  
\usepackage{graphicx} 
\usepackage{natbib}  
\usepackage{caption} 
\usepackage[linesnumbered,ruled,vlined]{algorithm2e}
\usepackage{amsmath}
\usepackage{amssymb}
\usepackage{enumitem}
\usepackage{tabularx}
\usepackage{booktabs}
\usepackage{xcolor}

\nocopyright

\usepackage{newfloat}
\usepackage{listings}
\DeclareCaptionStyle{ruled}{labelfont=normalfont,labelsep=colon,strut=off} 
\title{DualOpt: A Dual Divide-and-Optimize Algorithm for the Large-scale Traveling Salesman Problem}
\author{
    Shipei Zhou\textsuperscript{\rm 1}\equalcontrib,
    Yuandong Ding\textsuperscript{\rm 1}\equalcontrib,
    Chi Zhang\textsuperscript{\rm 1},
    Zhiguang Cao\textsuperscript{\rm 2},
    Yan Jin\textsuperscript{\rm 1}\thanks{Yan Jin is the corresponding author: jinyan@mail.hust.edu.cn}
}
\affiliations{
    \textsuperscript{\rm 1}School of Computer Science, Huazhong University of Science and Technology, China\\
    \textsuperscript{\rm 2}{School of Computing and Information Systems, Singapore Management University, Singapore}\\

}

\usepackage{bibentry}

\begin{document}
\maketitle

\begin{abstract}
This paper proposes a dual divide-and-optimize algorithm (DualOpt) for solving the large-scale traveling salesman problem (TSP). DualOpt combines two complementary strategies to improve both solution quality and computational efficiency. The first strategy is a grid-based divide-and-conquer procedure that partitions the TSP into smaller sub-problems, solving them in parallel and iteratively refining the solution by merging nodes and partial routes. The process continues until only one grid remains, yielding a high-quality initial solution. The second strategy involves a path-based divide-and-optimize procedure that further optimizes the solution by dividing it into sub-paths, optimizing each using a neural solver, and merging them back to progressively improve the overall solution. Extensive experiments conducted on two groups of TSP benchmark instances, including randomly generated instances with up to 100,000 nodes and real-world datasets from TSPLIB, demonstrate the effectiveness of DualOpt. The proposed DualOpt achieves highly competitive results compared to 10 state-of-the-art algorithms in the literature. In particular, DualOpt achieves an improvement gap up to 1.40\% for the largest instance TSP100K with a remarkable 104x speed-up over the leading heuristic solver LKH3. Additionally, DualOpt demonstrates strong generalization on TSPLIB benchmarks, confirming its capability to tackle diverse real-world TSP applications.

\end{abstract}

%

\section{Introduction}

The Traveling Salesman Problem (TSP) is an NP-hard combinatorial optimization problem, which has numerous real-world applications \cite{madani2021balancing,hacizade2018ga,matai2010traveling}. Let $G = (V, E)$ represent an undirected graph, where $V = \{v_{i} \mid 1 \leq i \leq N \}$ is the set of nodes and $E = \{e_{ij} \mid 1 \leq i, j \leq N \}$ is the set of edges, with $N$ being the total number of nodes. For each edge $e_{ij}$, the travel cost $cost(i, j)$ is defined as the Euclidean distance between the nodes $v_{i}$ and $v_{j}$. A special node $v_{d} \in V$ serves as the depot, from which the salesman starts and ends the trip. A feasible solution of TSP is defined as a Hamiltonian cycle that starts from the depot, visits each node exactly once, and ends at the depot. The objective is to minimize the total travel cost of the solution route $\tau$, denoted as $L(\tau) = \displaystyle\sum_{i=1}^{N-1}cost(\tau_{i}, \tau_{i+1}) + cost(\tau_{N}, \tau_{1})$, where $\tau_{i}$ represents the $i-{th}$ node in the route.

 Due to its theoretical and practical interest, various traditional exact/heuristic and machine learning-based algorithms have been proposed in the literature. While exact algorithms are generally computationally infeasible for large-scale instances, heuristics can provide near-optimal solutions for TSP with millions of cities, though they cannot guarantee optimality. They often involve time-consuming iterative searches, making them less suitable for time-sensitive tasks. Machine learning based algorithms, on the other hand, offer high computational efficiency and can achieve solution quality comparable to traditional methods for small-scale TSP instances. However, applying them to large-scale TSP, especially those with over 1,000 cities, remains a challenge. One approach is to leverage pre-trained models from small-scale instances for larger ones, but this often results in poor performance due to distribution shift~\cite{joshi2022learning}. Training models specifically for large-scale TSP is also impractical due to computational resource limitations.

A basic approach to deal with the large-scale routing problem is to apply the general principle of ``divide-and-conquer" {\cite{fu2021generalize,pan2023h,hou2023generalize,chen4679437extnco,ye2024glop, zheng2024udc,xia2024position}. It involves decomposing TSP into a subset of small sub-problems, which can be efficiently solved using traditional or machine learning algorithms to attain sub-solutions. The final solution is then obtained by combining these sub-solutions. This approach can significantly reduce the computational complexity of the large-scale TSP, and yield high-quality solutions in a relatively short time. 

In this work, we introduce a novel dual divide-and-optimize algorithm DualOpt based on the divide-and-conquer framework. The first divide-and-optimize procedure partitions the nodes into $M \times M$ equal-sized grids based on their coordinates. The nodes within each grid are initially solved using the well-known LKH3 solver \cite{helsgaun2017extension}. Next, an edge-breaking strategy is proposed to decompose the route into partial routes and nodes, significantly reducing the number of nodes. Subsequently, every four adjacent grids are merged into one larger grid. This process is repeated until all nodes are contained within a single grid, at which point the LKH3 solver is applied to obtain a complete route based on the partial routes and nodes. The second divide-and-optimize procedure further refines the obtained route by partitioning it into non-overlapping subpaths. These subpaths are then optimized in parallel using a neural solver \cite{kim2021learning,pan2023h, ye2024glop}, ultimately leading to an optimized route. The main contributions of this work are summarized as follows:
\begin{itemize}
\item We propose the DualOpt algorithm, which combines two complementary divide-and-conquer frameworks to address the large-scale TSP. The first framework employs grid-based partitioning, while the second applies path-based optimization, significantly improving both solution quality and computational efficiency of large-scale TSP.
\item We introduce a novel edge-breaking strategy that decomposes routes into partial routes and nodes. By representing these partial routes solely by their start and end nodes, this strategy considerably reduces the number of nodes, thereby decreasing computational complexity and improving search efficiency.
\item We conduct extensive experiments on both randomly generated and real-world large-scale TSP instances. To the best of our knowledge, DualOpt achieves state-of-the-art performance compared to other machine learning-based approaches, particularly excelling in large-scale instances with up to 100,000 nodes. 
\end{itemize}

\section{Related Work}
Here we present representative traditional and machine learning-based algorithms to solve TSP, and then focus on the divide-and-conquer algorithms that are more effective for solving large-scale TSP over 1000 nodes.
\paragraph{Traditional Algorithms}
Traditional algorithms can be roughly classified into two categories: exact and heuristic algorithms ~\cite{gutin2006traveling, accorsi2021fast}. Concorde ~\cite{applegate2009certification} is one of the best exact solvers, which models TSP as a mixed-integer programming problem and solves it using a branch-and-cut ~\cite{toth2002vehicle}. Exact algorithms can theoretically guarantee optimal solutions for instances of limited size, but are impractical for solving large instances due to their inherent exponential complexity. LKH3 ~\cite{helsgaun2017extension} is one of the state-of-the-art heuristics that uses the $k$-opt operators to find neighboring solutions, which is guided by a $\alpha$-nearness measure based on the minimum spanning tree. The heuristics are the most widely used algorithms in practice, yet they are still time consuming to obtain high-quality solutions when solving problems with tens of thousands of nodes.

\paragraph{Machine Learning-based Algorithms}
In recent years, machine learning-based algorithms have attracted more interest in solving the TSP. Depending on how solutions are constructed, they can be broadly categorized into \emph{end-to-end} approaches and \emph{search-based} approaches.

\emph{End-to-end} approaches generate solutions from scratch. The Attention Model \cite{kool2018attention} utilizes the Transformer architecture \cite{vaswani2017attention} and is trained with REINFORCE \cite{wiering2012reinforcement} using a greedy rollout baseline. POMO \cite{kwon2020pomo} improves on this by selecting multiple nodes as starting points, using symmetries in solution representation, and employing a shared baseline to enhance REINFORCE training. DIFUSCO \cite{sun2024difusco} uses graph-based denoising diffusion models to generate solutions, which are further optimized by local search with $k$-opt operators. Although DIFUSCO can handle problems with up to 10,000 nodes, it is less suitable for time-sensitive scenarios. Pointerformer \cite{jin2023pointerformer} incorporates a reversible residual network in the encoder and a multi-pointer network in the decoder, allowing it to solve problems with up to 1000 nodes. \emph{Search-based} approaches start with a feasible solution and iteratively apply predefined rules to improve it. NeuRewriter \cite{chen2019learning} iteratively rewrites local components through a region-picking and rule-picking process, with the model trained using Advantage Actor-Critic, and the reduced cost per iteration serves as the reward. NeuroLKH \cite{xin2021neurolkh} enhances the traditional LKH3 solver by employing a sparse graph network trained through supervised learning to simultaneously generate edge scores and node penalties, which guide the improvement process. DeepACO \cite{ye2024deepaco} integrates neural enhancements into Ant Colony Optimization (ACO) algorithms, further improving its performance.

\paragraph{Divide-and-Conquer Algorithms}}

Machine learning-based algorithms perform well on TSP instances with up to 1,000 nodes, but struggle with larger instances due to the exponential increase in memory requirements and computation time as the number of nodes grows. This leads to memory and time constraints during training, making it difficult to converge to near-optimal solutions. To address this challenge in solving large-scale problems with thousands or more nodes, the divide-and-conquer strategy is often employed. It is always combined with traditional or learning-based algorithms to generate high-quality solutions in a relatively short time. GCN-MCTS \cite{fu2021generalize} applies graph sampling to construct fixed-size sub-problems, solved by graph convolutional networks, with heatmaps guiding Monte-Carlo Tree Search (MCTS). H-TSP \cite{pan2023h} hierarchically builds TSP solutions using a two-level policy: the upper-level selects sub-problems, while the lower-level generates and merges open-loop routes. ExtNCO \cite{chen4679437extnco} uses LocKMeans with $o(n)$ complexity to divide nodes into sub-problems, solving them with neural combinatorial optimization and merging solutions via a minimum spanning tree. GLOP \cite{ye2024glop} learns global partition heatmaps to decompose large-scale problems and introduces a scalable real-time solver for small Shortest Hamiltonian Path problems. UDC \cite{zhengudc} proposes a Divide-Conquer-Reunion framework using efficient Graph Neural Networks for division and fixed-length solvers for sub-problems. SoftDist \cite{xia2024position} demonstrates that a simple baseline method outperforms complex machine learning approaches in heatmap generation, and the heatmap-guided MCTS paradigm is inferior to the LKH3 heuristic despite leveraging hand-crafted strategies.

\section{The Proposed DualOpt Algorithm}
The proposed DualOpt algorithm is based on and extends the basic divide-and-conquer method by incorporating a grid-based procedure and a path-based procedure to improve efficiency. Each procedure operates on two levels: the first level is responsible for generating sub-problems, while the second level focuses on solving these sub-problems.
\begin{algorithm}[h]\small
    \caption{The Proposed DualOpt Algorithm for the Large-scale TSP}\label{alg:DualOpt}
    \KwIn{TSP instance $V=\{v_1, v_2, \cdots, v_N \}$}
    \KwOut{Solution route $\tau$}
    $PartialRoutesSet$ $\Upsilon \gets \emptyset$ \;
    $NodesSet$ $\aleph \gets V$ \;    
    \Repeat{predefined iterations reached}{
    $Grids \gets Partition(\Upsilon,\aleph)$ \;
    $Routes \gets SolveGridsParallel(\Upsilon, \aleph)$ \;
    $\Upsilon', \aleph' \gets EdgeBreaking(Routes) $ \;  
    $\Upsilon \gets \Upsilon' $ \;  
    $\aleph \gets \aleph' $ \; 
    }
    $\tau \gets SolveReducedProb(\Upsilon, \aleph) $ \;
    \Repeat{predefined iterations reached}{
    $SubPath \gets DivideSolution(\tau) $\;
    $SubPath' \gets OptimizeSubPathBatch(SubPath) $\;    
    $\tau' \gets MergeSubPath(SubPath')$ \;
    $\tau \gets \tau'$ \;
    }
    return $\tau$
\end{algorithm}

The whole procedure of DualOpt is summarized in Algorithm \ref{alg:DualOpt}. It starts by partitioning the TSP instance into smaller grids, which reduces the problem size and allows for parallel solving. In each iteration, the partial routes and nodes within each grid are solved in parallel to generate a solution for that portion of the TSP. Next, an edge-breaking procedure is applied to divide each route into smaller partial routes and nodes. The grids are then merged into larger ones using a grid merging procedure. This grid-based iteration continues until only one grid remains, forming a reduced problem that consists of a number of partial routes and nodes. From this reduced problem, a high-quality initial solution of is constructed. To further refine the solution, it is divided into sub-paths, which are optimized in batches and then merged. This iterative refinement, performed with varying partition sizes, progressively improves the solution quality.

\subsection{A Grid-based Divide-and-Conquer Procedure}
To decompose the large-scale TSP for efficient solving without significantly downgrading solution quality, we employ an iterative grid decomposition strategy, denoted as the Grid-based Divide-and-Conquer Procedure, as depicted in Algorithm \ref{alg:Gridbased}. Initially, the node set \(\aleph\) contains all the nodes of the TSP instance, while the partial route set \(\Upsilon\) is initialized as empty. In each iteration, the 2D space is discretized into an evenly spaced grid of size \(K = 2^{N_{iter}-iter} \times 2^{N_{iter}-iter}\) grids. The node set \(\aleph\) and partial route set \(\Upsilon\) are then divided into $K$ subsets \(\{(\Upsilon_{1}, \aleph_{1}), \ldots, (\Upsilon_{K}, \aleph_{K})\}\) based on their positions within the grid. Each of these subsets is solved in parallel using the well-known TSP solver LKH3 \cite{helsgaun2017extension}, resulting in $K$ small routes \(\{R_{1}, \ldots, R_{K}\}\). If the current iteration (\emph{iter}) is not the last one, the algorithm proceeds by applying a proposed edge-breaking strategy in parallel, which breaks a subset of the edges of the routes and updates \(\Upsilon\) and \(\aleph\) with new sets of partial routes and nodes \(\{(\Upsilon'_{1}, \aleph'_{1}), \ldots, (\Upsilon'_{K}, \aleph'_{K})\}\). As the iterations progress, $K$ decreases and the grid size increases correspondingly until only one grid remains. At this final stage, a reduced problem consisting of partial routes and nodes is formed, from which an initial high-quality TSP solution \(\tau\) is obtained.
\begin{algorithm}[!htp]\small
    \caption{Grid-based Divide-and-Conquer }\label{alg:Gridbased}
    \KwIn{TSP instance \emph{$V = \{v_{1},v_{2},...,v_{N}\}$}, number of iterations \emph{$N_{iter}$}}
    \KwOut{Solution route \emph{$\tau$}}
    
  $iter \gets 1$ \;
  $NodesSet$ $\aleph \gets V$ \;
  $PartialRoutesSet$ $\Upsilon \gets \emptyset$ \;
    
  \While{$iter\le N_{iter}$}{
    $K \gets 2^{N_{iter}-iter}\times 2^{N_{iter}-iter}$; /*Calculate the number of grids*/ 
    
    $\{(\Upsilon_{1},\aleph_{1}),\ldots,(\Upsilon_{K},\aleph_{K})\} \gets Partition(\Upsilon, \aleph, K)$;  /*Partition nodes and partial routes into grids*/ 

    $\{R_{1},...,R_{K}\} \gets SolveGridsParallel( \{(\Upsilon_{1},\aleph_{1}),\ldots,(\Upsilon_{K},\aleph_{K})\}$);  /*Solve the TSP for each grid in parallel*/ 

    \If{$iter \neq N_{iter}$}{
        $\{(\Upsilon'_{1},\aleph'_{1}),\ldots,(\Upsilon'_{K},\aleph'_{K})\} \gets BreakEdgesParallel(\{R_{1},\ldots,R_{K}\}$; /*Break edges into partial routes and nodes in parallel*/
        
        $\Upsilon$ is updated by $\{\Upsilon'_{1}, \ldots,\Upsilon'_{K}\}$ \;  
        $\aleph$ is updated by $\{\aleph'_{1}, \ldots,\aleph'_{K}\}$ \; 
        
    } \Else{
        $\tau \gets SolveReducedProb(\Upsilon, \aleph)$ ; /*Solve the reduced problem to get a complete route*/
    }
    $iter \gets iter + 1$ \;
  }
  return $\tau$
\end{algorithm}

\begin{figure}[!tp]
  \centering
  \includegraphics[scale=0.05]{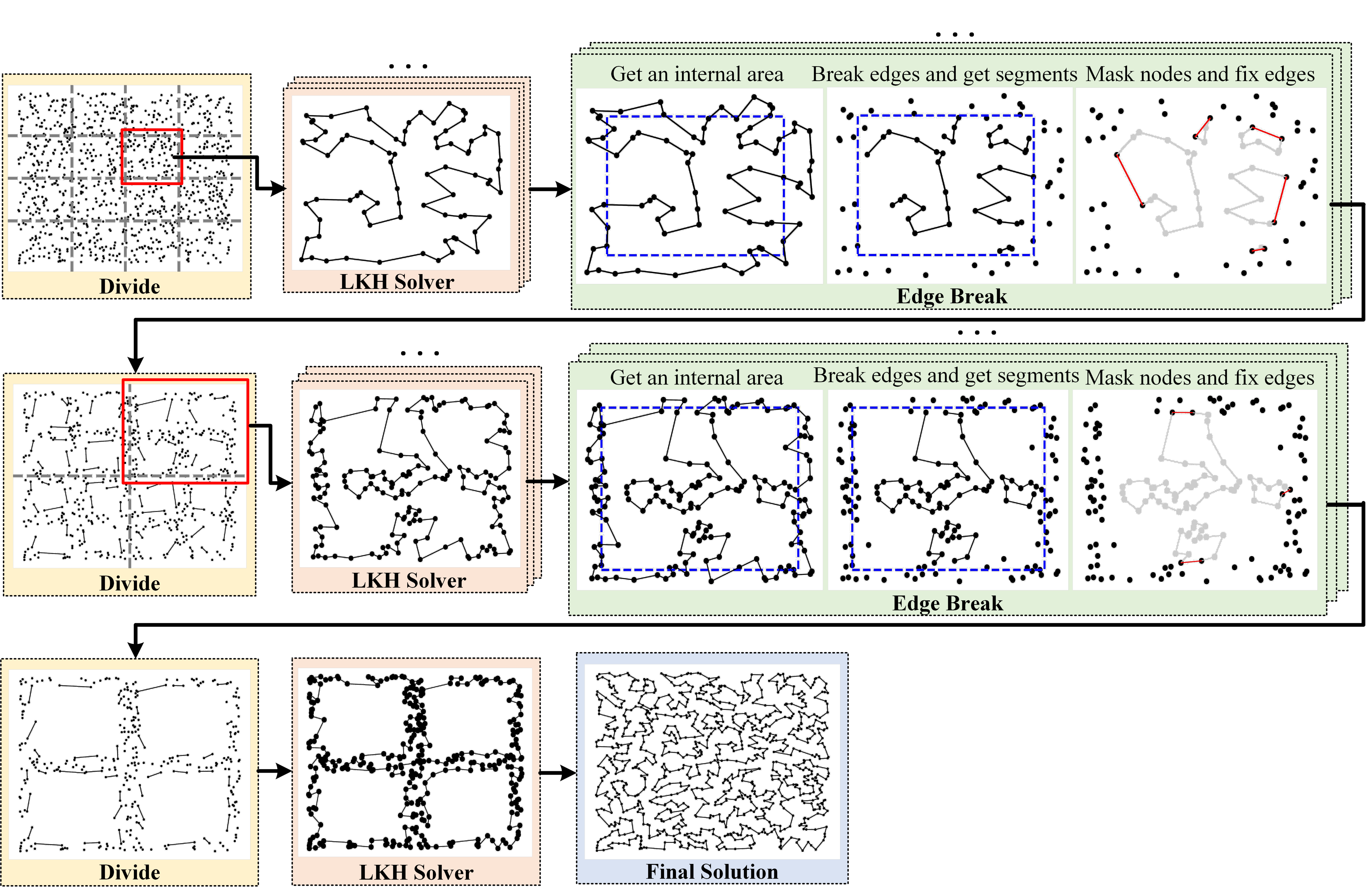}
  \caption{An illustration of the Grid-based Divide-and-Conquer Procedure with $N_{iter} = 3$.}
  \label{fig:example_grid}
\end{figure}

Figure \ref{fig:example_grid} illustrates this procedure with $N_{iter} = 3$. In the first iteration, $K$ is 16, partitioning the 2D space into 16 evenly spaced grids. The LKH3 generates 16 localized routes per grid, which are then decomposed into partial routes and separate nodes. In the second iteration, $K$ is reduced to 4, resulting in 4 larger grids where the LKH3 processes partial routes with fixed nodes. In the final iteration, $K$ is reduced to 1, merging the grids into a single one. As a result, the number of nodes decreases while the number of partial routes increases. This reduction in problem scale enables the LKH3 solver to efficiently find a TSP solution with high quality.

\textbf{The LKH3 Solver} The LKH3 solver, a state-of-the-art heuristic for the TSP, is built upon the foundational Lin-Kernighan heuristic, incorporating several advanced components. Among these, the use of a 1-tree structure is particularly notable, as it provides a lower bound that guides the search process effectively. The algorithm dynamically employs $K$-opt neighborhood operators, where multiple edges are iteratively removed and reconnected to explore potential routes within the solution space. The efficiency of LKH3 is further enhanced by using candidate sets, which strategically limits the number of neighborhood operators, thereby reducing search overhead. 

The number of nodes within each grid, typically ranging from a few hundreds to approximately one thousand, is determined by both the instance scale and the number of grids $K$. The input to the solver includes not only the coordinates of nodes but also the fixed partial routes. LKH3 excels in handling node scales within this range, delivering high-quality solutions with remarkable speed, particularly for routing problems involving fixed edges. Therefore, LKH3 was selected as the subsolver for this study.

\textbf{Edge-breaking Strategy} Recall that multiple routes are generated, one for each grid within the 2D space. When merging and optimizing these routes, it is essential to reschedule the surrounding nodes of each grid in conjunction with the nodes of adjacent grids, while excluding the internal nodes from this rescheduling process. To achieve this, we propose an edge-breaking strategy that effectively reduces the problem's scale. Specifically, for each grid, we define an internal grid that maintains a fixed spacing from the outer grid. 
This spacing is calculated as \text{spacing} = $\frac{x_{\text{max}} - x_{\text{min}}}{2^{N_{\text{iter}} + 2}}$, where \( x_{\text{max}} \) and \( x_{\text{min}} \) denote the maximum and minimum $x$-coordinates of the grid, respectively. 
We then remove all edges that lie outside the internal grid, as well as those crossing into it. As a result, the nodes and edges within the internal grid form several partial routes that each may consist of more than two connected nodes, along with isolated nodes.

\subsection{A Path-based Divide-and-Optimize Procedure}
Based on the solution derived from the grid-based divide-and-conquer procedure, further optimization is achieved through the path-based divide-and-optimize procedure, as detailed in Algorithm \ref{alg:Pathbased}. Following previous studies~\cite{kim2021learning,ye2024glop,zhengudc}, this iterative algorithm aims to refine an initial solution $\tau$ by partitioning it into smaller sub-paths, optimizing these sub-paths individually, and then merging them to generate an improved solution $\tau^*$. 

The procedure begins by iterating over a specified set of sub-path lengths ${len_1, \ldots, len_m}$ and iterations ${iter_{1}, \ldots, iter_{m}}$. For each sub-path length $len$, the initial solution $\tau$ is divided into sub-paths of the designated length. If the length of the last sub-path is shorter than the designated length, it remains unchanged. Each sub-path can be considered as an open-loop TSP with two fixed endpoints. These sub-paths undergo a distribution normalization of vertex coordinates to maintain consistency with the original TSP instance, resulting in normalized sub-paths denoted as $SubPathSet^{'}$. 
Following normalization, a neural solver is employed in batch mode to optimize the normalized sub-paths, and sub-paths are updated by the policy when they are better than the current ones, producing a set of sub-paths $SubPathSet^{*}$. These sub-paths are then merged to form an improved solution $\tau^*$ by connecting their fixed endpoints in their original order.
To ensure continuous refinement and overlap during the iterations, the starting point $\kappa$ for sub-path division is incremented by $\max(1, \frac{len}{iter})$. This iterative refinement process is repeated for each combination of sub-path length and iteration count, progressively refining the manageable sub-paths of the initial solution $\tau$ to yield a highly optimized solution $\tau^*$.
\begin{algorithm}[h]\small
    \caption{Path-based Divide-and-Optimize}\label{alg:Pathbased}
    \KwIn{Solution $\tau = \{\tau_{1}, \tau_{2},...,\tau_{N}\}$, a set of sub-path lengths$\{len_1, \ldots, len_m\}$, and a set of iterations $\{iter_{1}, \ldots , iter_{m} \}$}
    \KwOut{Optimized Solution \emph{$\tau^*$}}
    
  \For{$len \gets len_1$ \textbf{to} $len_{m}$}{
    $\kappa  \gets 1$;

     \For{$iter \gets iter_1 $ \textbf{to} $ iter_{m}$} {
     $SubPathSet \gets \{\tau_{\kappa : \kappa +len}, \tau_{\kappa +len: \kappa +2len}, \ldots \}$;

     $SubPathSet^{'} \gets NormalizeVertexCoordinate(SubPathSet$);
     
     $SubPathSet^* \gets NeuralSolverBatch(SubPathSet^{'}$);
     
     $\tau^* \gets MergeSubPath(SubPathSet^*$);
     
     $\kappa  \gets \kappa  + max(1, \frac{len}{iter})$;
     
     $\tau \gets \tau^*$
     }
    }
  return $\tau^*$
\end{algorithm}

\textbf{Neural Solver}
The underlying solver of the path-based divide-and-optimize procedure is an attention-based neural network, which can efficiently solve the obtained sub-paths through batch parallelism. 

The neural network uses an encoder-decoder architecture. The encoder employs self-attention layers to embed the input node sequence, while the decoder generates the node sequence auto-regressively. Each sub-path $\pi$ can be regarded as an open-loop TSP with specified start and end nodes. We employ a modified context embedding inspired by \cite{kim2021learning}, defined as ${h}_{(c)}^{(L)}=[\overline{h}^{(L)}, {h}_{{\pi}_{pre}}^{(L)},{h}_{{\pi}_{des}}^{(L)}]$. Here, $h$ denotes a high-dimensional embedding vector from the encoder, and $L$ represents the number of multi-head attention layers. $\overline{h}^{(L)}$ is the mean of the final node embeddings, ${h}_{{\pi}_{pre}}^{(L)}$ is the embedding of the previously selected node, and ${h}_{{\pi}_{des}}^{(L)}$ is embedding of the end node of the sub-path.
Moreover, following insights from previous studies ~\cite{kim2021learning,cheng2023select,ye2024glop}, we leverage the symmetry of sub-paths. Specifically, reversing the traversal direction of a sub-path (moving backward versus forward) yields an equivalent sequence, which helps improve the network's robustness and efficiency.



The network employs a single-step constructive policy aimed at optimizing the solution by generating sequences in both forward and backward directions. The policy is defined as follows, where $s$ represents a sub-path with a start node ${\pi}_{1}$ and an end node ${\pi}_{n}$. The policy ${p}_{\theta}(\boldsymbol{{\pi}_{f}}, \boldsymbol{{\pi}_{b}}|s)$ is parameterized by $\theta$ and specifies a stochastic policy for constructing the forward solution $\boldsymbol{{\pi}_{f}}$ and the backward solution $\boldsymbol{{\pi}_{b}}$:
\begin{equation}
\begin{aligned}
    &{p}_{\theta}(\boldsymbol{{\pi}_{f}}, \boldsymbol{{\pi}_{b}}|s) = {p}_{\theta}(\boldsymbol{{\pi}_{f}}|s){p}_{\theta}(\boldsymbol{{\pi}_{b}}|s)\\
    &=\displaystyle\prod_{t=1}^{n-1}{p}_{\theta}({\pi}_{1+t}|s, \boldsymbol{{\pi}_{1:t}},{\pi}_{n}) \times \displaystyle\prod_{t=1}^{n-2}{p}_{\theta}({\pi}_{n-t}|s, \boldsymbol{{\pi}_{n:n-t+1}},{\pi}_{1}).
\end{aligned}
\end{equation}

The model is trained using REINFORCE with a shared baseline. The objective is to minimize the expected length of the solutions to the sub-problems. To achieve this, the loss function is defined as the expected length of these solutions. The policy gradient is then computed as follows:
\begin{equation}
    \begin{aligned}
    \triangledown \mathcal{L}(\theta|s)=&\mathrm{E}_{p_{\theta}(\boldsymbol{{\pi}_{f}}|s)}[R(\boldsymbol{{\pi}_{f}})-b(s)]\triangledown \log p_{\theta}(\boldsymbol{{\pi}_{f}}|s)\\
    &+ \mathrm{E}_{p_{\theta}(\boldsymbol{{\pi}_{b}}|s)}[R(\boldsymbol{{\pi}_{b}})-b(s)]\triangledown \log p_{\theta}(\boldsymbol{{\pi}_{b}}|s).
    \end{aligned}
\end{equation}

The shared baseline $b(s)$ is employed to reduce variance and enhance training stability. This baseline is computed by averaging the path lengths obtained from two greedy rollouts. Note that during model training, both forward and backward solutions are utilized. However, during inference, only the superior trajectory between the two solutions is selected to ensure the best possible outcome.

\textbf{Node Coordinate Normalization}
To improve the robustness of the model with respect to the sub-paths, we normalize the node coordinates to be uniformly distributed within the range $[0, 1]$ \cite{fu2021generalize}. We define $x_{max} = \max_{i\in G}{x_{i}}$, $x_{min} = \min_{i\in G}{x_{i}}$, $y_{max} = \max_{i\in G}{y_{i}}$, $y_{min} = \min_{i\in G}{y_{i}}$ as the maximum and minimum values of the horizontal and vertical coordinates of all $N$ nodes in the TSP instance. For each node in the sub-path, we convert its coordinate $(x_{i}, y_{i})$ to $(x_{i}^{'}, y_{i}^{'})$ as shown in Eq. (\ref{eq:normalization}), ensuring that all coordinates fall within the range $[0,1]$.
\begin{equation}
    \begin{split}
        x_{i}^{'} &= \frac{x_{i} - x_{\min}}{\max(x_{\max} - x_{\min}, y_{\max} - y_{\min})}, \\
        y_{i}^{'} &= \frac{y_{i} - y_{\min}}{\max(x_{\max} - x_{\min}, y_{\max} - y_{\min})}.
    \end{split} \label{eq:normalization}
\end{equation}

\section{Experiments}
To evaluate the performance of our proposed DualOpt, we conducted extensive experiments on the large-scale TSP instances, with up to 100,000 nodes. We compared the performance of DualOpt with that of state-of-the-art algorithms from the literature. All experiments, including rerunning the baseline algorithms, were executed on a machine with an RTX 3080(10GB) GPU and a 12-core Intel(R) Xeon(R) Platinum 8255C CPU.

Our evaluation focused on two groups of the large-scale instances: randomly generated instances \textbf{TSP\_random} and the well-known real-world benchmark \textbf{TSPLIB}. The \textbf{TSP\_random} represents random Euclidean instances, with node coordinates uniformly sampled from a unit square $[0,1]^{2}$. This group includes seven datasets, each corresponding to a different number of nodes, denoted as TSP$N$, i.e., TSP1K (1,000 nodes), TSP2K, TSP5K, TSP10K, TSP20K, TSP50K, and TSP100K (100,000 nodes). For a fair comparison, we used the same test instances provided by Fu et al. \cite{fu2021generalize} and Pan et al. \cite{pan2023h}. Each dataset contains 16 instances, except TSP1K, which includes 128 instances, and TSP100K, which contains one instance. The \textbf{TSPLIB} is a well-known real-world benchmark \cite{reinelt1991tsplib} that consists of 100 instances with diverse node distributions derived from practical applications, with sizes ranging from 14 to 85,900 nodes. For our experiments, we select the ten largest TSPLIB instances, each containing more than 5,000 nodes.

\paragraph{Baselines} 
We compare DualOpt against ten leading TSP algorithms in the literature. \textbf{Traditional Algorithms:} 1) Concorde \cite{cook2011traveling}: One of the best exact solvers. 2) LKH3 \cite{helsgaun2017extension}: One of the highly optimized heuristic solvers. \textbf{Machine Learning Based Algorithms:} 1) POMO \cite{kwon2020pomo}: Feature an \emph{end-to-end} model based on Attention Model, comparable to LKH3 for small-scale TSP instances. 2) DIMES \cite{qiu2022dimes}: Introduce a compact continuous space for parameterizing the underlying distribution of candidate solutions for solving large-scale TSP with up to 10K nodes. 3) DIFUSO \cite{sun2024difusco}: Leverage a graph-based denoising diffusion model to solve large-scale TSP with up to 10K nodes. 4) SIL \cite{luo2024self}: Develop an efficient self-improved mechanism that enables direct model training on large-scale problem instances without labeled data, handling TSP instances with up to 100K nodes. \textbf{Divide-and-Conquer Algorithms:} 1) GCN+MCTS \cite{fu2021generalize}: Combine a GCN model trained with supervised learning and MCTS to solve large-scale TSP involving up to 10K nodes with a long searching time. 2) SoftDist \cite{xia2024position}: Critically evaluate  machine learning guided heatmap generation, the heatmap-guided MCTS paradigm for large-scale TSP. 3) H-TSP \cite{pan2023h}: Hierarchically construct a solution of a TSP instance with up to 10K nodes based on two-level policies. 4) GLOP \cite{ye2024glop}: Decompose large routing problems into Shortest Hamiltonian Path Problems, further improved by local policy. It is the first neural solver to effectively scale to TSP with up to 100K nodes.

\paragraph{Parameter Settings}
For the grid-based divide-and-conquer procedure, we set the value of $N_{iter}$ for different TSP problem scales, as shown in Table \ref{Setting of N_{iter}}. The LKH3 solver is used with its default parameters as specified in the literature \cite{helsgaun2017extension} both in DualOpt and baseline. LKH3 baseline runs instance-by-instance. In the grid-based divide-and-conquer stage, LKH3 sub-solver just run parallel for grids in a single instance. For the path-based divide-and-optimize procedure, we train different neural solvers with graph size $len=\{50, 20, 10\}$, with node coordinates randomly generated from a uniform distribution within a unit square. During the training process, we use the same hyper-parameter settings as those used by Kool et al \cite{kool2018attention}. And during test time, we set $iter = \{25, 10, 5\}$.  Our code is publicly available.\footnotemark[1]

\footnotetext[1]{https://github.com/Learning4Optimization-HUST/DualOpt}
\begin{table}[!h]
        \centering
        \footnotesize
        \renewcommand\arraystretch{0.9}
        \begin{tabularx}{\columnwidth}{XX} 
            \toprule 
             Number of Iterations & Problem Scale of TSP  \\
            \midrule 
            $N_{iter} = 2$ & $N < 5,000$\\
            $N_{iter} = 3$ & $5,000 \leq N < 20,000$\\
            $N_{iter} = 4$ & $20,000 \leq N < 100,000$\\
            $N_{iter} = 5$ & $N \geq 100,000$\\
            \bottomrule
        \end{tabularx}
        \caption{Parameter setting of $N_{iter}$.}
        \label{Setting of N_{iter}}
\end{table}

\paragraph{Comparative Studies}
\begin{table*}[t]
  \centering
  \footnotesize
  \renewcommand\arraystretch{0.9}
  \begin{tabularx}{\textwidth}{c|*{3}{X}| *{3}{X}| *{3}{X}}
    \specialrule{.15em}{.1em}{.2em}
    & &TSP1K& & &TSP2K& & &TSP5K&\\
    Algorithm & Obj. &Gap(\%) & Time & Obj. &Gap(\%) & Time & Obj. &Gap(\%) & Time\\
    \midrule
    LKH3 & 23.31 & 0.00 & 2.4s & 32.89 & 0.00 & 9.3s & 51.52 & 0.00 & 1.3m \\
    Concorde & 23.12 & -0.82 & 7m & 32.44 & -1.37 & 2.8h & 53.45 & 3.75 & 3.2h\\
    \midrule
    POMO$\dagger$ & 30.52 & 30.93 & 4.3s & 46.49 & 41.35 & 35.9s & 80.79 & 56.81 & 9.6m\\
    DIMES$\dagger$ & 23.69 & 1.63 & 2.2m & - & - & - & - & - & -\\
    DIFUSCO$\dagger$ & 23.39 & 0.34 & 11.5s & - & - & - & - & - & - \\
    SIL$\dagger$ & \textbf{23.31} & 0.00 & 0.6s & - & - & - & 51.92 & 0.78 & 28.5s \\
     \midrule
    GCN-MCTS$\dagger$ & 23.86 & 2.36 & 5.8s & 33.42 & 1.61 & 3.3m & 52.83 & 2.54 & 6.3m \\
    SOFTDIST$\dagger$ & 23.63 & 1.37 & 1.57 & - & - & - & - & - & -\\
    H-TSP & 24.66 & 5.79 & 0.7s & 35.22 & 7.08 & 1.5s & 55.72 & 8.15 & 2.3s\\
    GLOP & 24.01 & 3.00 & 0.4s & 33.90 & 3.07 & 0.9s & 53.49 & 3.82 & 1.8s \\
    \midrule
     DualOpt  & \textbf{23.31} & \textbf{0.00} & 5.6s & \textbf{32.72} & \textbf{-0.52} & 7.6s & 51.56 & 0.08 & 13.9s \\
    \specialrule{.15em}{.2em}{.0em}
  \end{tabularx}

  \begin{tabularx}{\textwidth}{>{\centering\arraybackslash}p{1.5cm}|*{3}{X}| *{3}{X}| *{3}{X}|*{3}{X}}
    \specialrule{.15em}{.1em}{.2em} 
    & &TSP10K& & &TSP20K& & &TSP50K& & &TSP100K& \\
    Method & Obj. &Gap(\%) & Time & Obj. &Gap(\%) & Time & Obj. &Gap(\%) & Time & Obj. &Gap(\%) & Time\\
    \midrule
    LKH3 & 72.96 & 0.00 & 6.3m & 103.28 & 0.00 & 27.4m & 164.08 & 0.00 & 3.0h & 234.098 & 0.00 & 14.9h \\
     \midrule
    DIMES$\dagger$ & 74.06 & 1.50 & 13m & - & - & - & - & - & - & - & - & - \\
    DIFUSCO$\dagger$ & 73.62 & 0.90 & 3m & - & - & - & - & - & - & - & - & - \\
    SIL$\dagger$ & 73.32 & 0.49 & 51s & - & - & - & 164.53 & 0.27 & 3.8m & 232.66 & -0.61 & 7.5m \\
     \midrule
    GCN-MCTS$\dagger$ & 74.93 & 2.70 & 6.5m & - & - & - & - & - & - & - & - & - \\
    SOFTDIST$\dagger$ & 74.03 & 1.40 & 1m & - & - & - & - & - & - & - & - & - \\
    H-TSP & 78.45 & 7.52 & 4.8s & 110.7 & 7.18 & 10.4s &  & OOM &  &  &OOM &  \\
    GLOP & 75.42 & 3.37 & 3.5s & 106.7 & 3.31 & 7.9s & 168.2 & 2.51 & 12.1s & 237.99 & 1.66 & 2.5m \\
    \midrule
     DualOpt & \textbf{72.62} & \textbf{-0.47} & 33.9s & \textbf{102.9} & \textbf{-0.37} & 1m & \textbf{162.81} & \textbf{-0.77} & 6.5m & \textbf{230.83} & \textbf{-1.40} & 8.6m \\
    \specialrule{.15em}{.2em}{.0em}
  \end{tabularx}
   \caption{Comparative results with 10 leading algorithms on large-scale \textit{TSP\_random} with up to 100K nodes. }
    \label{tab:TSPrandom}
\end{table*}
Table \ref{tab:TSPrandom} presents a comparison of 10 leading algorithms and our DualOpt algorithm on randomly distributed datasets \textbf{TSP\_random}. The ``Obj." column shows the average objective lengths of the routes obtained by each algorithm for each instance, while the ``Gap" column measures the percentage difference between the average route lengths attained by each algorithm and the LKH3, considered as the ground truth, i.e., $Gap = \frac{\textit{Obj. of Algo. - Obj of LKH3}}{\textit{obj. of LKH3}} \times 100\%$. The ``Time" column indicates the average time required to solve each instance. The $\dagger$ following the algorithm indicates that the results of this algorithm are drawn from the literature directly and ``-" means the results are not provided in the literature. The ``OOM” stands for out of CUDA memory with our platform.

From Table \ref{tab:TSPrandom}, we can make the following comments about the \textbf{TSP\_random} instances. First, the heuristic solver LKH3 provides the highest quality solutions for all instances except TSP1K and TSP2K, though it requires long time searching. DualOpt matches or improves upon LKH3's results for all instances except TSP5K, and achieves an improved result with an improvement gap up to 1.40\% for the largest instance TSP100K, with a remarkable 104x speed-up. Second, DualOpt clearly outperforms machine learning based algorithms in both solution quality and running time. POMO underperforms across all instances, primarily because it cannot be trained directly on large-scale instances, and models trained on small-scale instances fail to generalize effectively to large-scale instances. Compared to the current state-of-the-art SIL, our DualOpt surpasses SIL in all instances except for a tie on TSP1K. Third, DualOpt consistently outperforms all four divide-and-conquer algorithms in terms of the best objective result, achieving better results than GLOP for all instances. These observations highlight the superiority of DualOpt in both solution quality and computational efficiency. 


To demonstrate the solver’s generalization capabilities, we then directly applied the trained neural solver to the real-world dataset TSPLIB with varied distributions. Table \ref{tab:TSPLIB} presents the results for the 10 largest instances from TSPLIB. Despite the different node distributions in TSPLIB compared to those in TSP\_random, DualOpt demonstrates impressive generalization capabilities. It achieves highly competitive results compared to three leading large-scale TSP algorithms from the literature. This performance highlights DualOpt's effectiveness in handling various types of TSP instances and further underscores its versatility in real-world applications.

\begin{table*}[!t]
\fontsize{9pt}{10pt}\selectfont 
\centering
\renewcommand\arraystretch{0.9}
  \begin{tabularx}{\textwidth}{c|>{\hsize=0.8\hsize}X>{\hsize=0.6\hsize}X|>{\hsize=0.9\hsize}X>{\hsize=0.6\hsize}X>{\hsize=0.5\hsize}X|>{\hsize=0.9\hsize}X>{\hsize=0.6\hsize}X>{\hsize=0.5\hsize}X|>{\hsize=0.9\hsize}X>{\hsize=0.6\hsize}X>{\hsize=0.5\hsize}X}

    \specialrule{.15em}{.2em}{.2em}
    
    &\multicolumn{2}{c|}{LKH3} & \multicolumn{3}{c|}{H-TSP} & \multicolumn{3}{c|}{GLOP} & \multicolumn{3}{c}{DualOpt} \\
    Instance & Obj.& Time & Obj. & Gap(\%)& Time & Obj.&Gap(\%)& Time & Obj. &Gap(\%) & Time\\
    \midrule
    rl5915	&	572085	&	9.9m 	&	652336& 14.03	&	9s	&	628928&9.94 	&	16s	&	\textbf{579152} &1.24	&	14s	\\
rl5934	&	559712	&	9.9m 	&	642214&14.74 	&	9s	&	616629&10.17 	&	16s	&	\textbf{565912} &1.11	&	19s	\\
pla7397	&	23382264 	&	12.4m 	&	25494130&9.03	&	12s	&	24990688&6.88	&	16s	&	\textbf{23536550} &0.66	&	41s	\\
rl11849	&	929001	&	18.9m	&	1046963&11.2&	17s	&	1006378& 8.3	&	17s	&	\textbf{935904}	&0.74&	49s	\\
usa13509	&	20133724 	&	22.6m 	&	21923532&8.89	&	20s	&	21023604&4.42 	&	17s	&	\textbf{20248476} &0.57	&	1.3m	\\
brd14051	&	474149 	&	23.5m 	&	506211&6.76 	&	20s	&	491735& 3.71 	&	18s	&	\textbf{474559} & 0.09	&	1m	\\
d15112	&	1588550 	&	25.2m 	&	1696577&6.80 	&	22s	&	1648777& 3.79 	&	18s	&	\textbf{1589033} &0.03	&	1.2m	\\
d18512	&	652911 	&	31m 	&	694116&6.31	&	26s	&	676840&3.66	&	21s	&	\textbf{652457} &0.07	&	1.3m	\\
pla33810	&	67084217	&	56.4m 	&	\multicolumn{3}{c|}{OOM}		&	71934504&7.23 	&	33s	&	\textbf{68083048}&1.49	&	1.2m	\\
pla85900	&	148763746 	&	4.5h	&	\multicolumn{3}{c|}{OOM}		&153463696&3.15	&	1.7m	&	\textbf{151378243}&1.80	&	3.5m	\\

    \specialrule{.15em}{.2em}{.0em}
  \end{tabularx}
  \caption{Comparative results on 10 largest instances of \textit{TSPLIB}.}
   \label{tab:TSPLIB}
\end{table*}
\paragraph{An Ablation Study}
Our DualOpt integrates two important divide-and-conquer procedures to achieve superior performance. To evaluate the impact of each procedure, we conducted an ablation study with two distinct variants of DualOpt: DualOpt$_{w/oPath}$ and DualOpt$_{w/oGrid}$. DualOpt$_{w/oPath}$ employs only the grid-based divide-and-conquer procedure, while DualOpt$_{w/oGrid}$ generates the initial solution using a simple yet effective approach called random insertion, which is then improved using the path-based divide-and-optimize procedure. The results of this ablation study are summarized in Table \ref{tab:Ablation}. The comparative results clearly demonstrate the significance of both procedures within DualOpt. Specifically, the original DualOpt consistently outperforms the variants, underscoring the importance of combining both the grid-based and path-based procedures for achieving superior performance.
\begin{table*}[!h] 
  \centering
  \footnotesize
  \renewcommand\arraystretch{0.9}
  \begin{tabularx}{\textwidth}{c|*{3}{X}| *{3}{X}| *{3}{X}}
    \specialrule{.15em}{.1em}{.2em}
    &\multicolumn{3}{c|}{DualOpt} & \multicolumn{3}{c|}{DualOpt$_{w/oPath}$} & \multicolumn{3}{c}{DualOpt$_{w/oGrid}$} \\
    Instance & Obj. &Gap(\%) & Time & Obj. &Gap(\%) & Time & Obj. &Gap(\%) & Time\\
    \midrule
   TSP1K	&	23.31	&	0.00 	&	5.6s	&	23.34	&	0.13 	&	4.5s	&	24.56	&	 5.36	&	5.2s	\\
TSP2K	&	32.72	&	-0.52 	&	7.6s	&	32.75	&	-0.43	&	6.3s	&	34.3	&	4.29 	&	5.3s	\\
TSP5K	&	51.56	&	0.08 	&	13.9s	&	51.65	&	0.25	&	12.1s	&	54.03	&	 4.87	&	7.2s	\\
TSP10K	&	72.62	&	-0.47 	&	33.9s	&	72.85	&	-0.15	&	31.7s	&	76.25	&	 4.51	&	8.2s	\\
TSP20K	&	102.9	&	-0.37 	&	1m	&	103.4	&	0.12	&	59s	&	107.76	&	 4.34	&	8.7s	\\
TSP50K	&	162.81	&	-0.77	&	6.5m	&	164	&	 -0.05	&	6.4m	&	170.24	&	 3.75	&	31.3s	\\
TSP100K	&	230.83	&	-1.40 	&	8.6m	&	234.2	&	0.04	&	8.5m	&	240.82	&	 2.87	&	2.5m	\\
    \specialrule{.15em}{.2em}{.0em}
  \end{tabularx}
  \caption{An ablation study of DualOpt and its two variants on \textit{TSP\_random}}.
  \label{tab:Ablation}
\end{table*}

\section{Conclusion}
In this paper, we introduced a novel dual divide-and-optimize algorithm DualOpt to solve large-scale traveling salesman problem, which integrates two complementary strategies: a grid-based divide-and-conquer procedure and a path-based divide-and-optimize procedure. Through extensive experiments on randomly generated instances and real-world datasets, DualOpt consistently achieves highly competitive results with state-of-the-art TSP algorithms in terms of both solution quality and computational efficiency. Moreover, DualOpt is able to generalize across different TSP distributions, making it a versatile solver for tackling diverse real-world TSP applications. The success of DualOpt paves the way for future research into more sophisticated divide-and-conquer strategies and their application to different types of other combinatorial optimization problems.  

\section{Acknowledgement}
This work is supported by National Natural Science Foundation 62472189.

\bibliography{aaai25}

\end{document}